\documentclass[11pt,a4paper]{article}
\pdfoutput=1
\PassOptionsToPackage{breaklinks}{hyperref}
\usepackage[hyperref]{acl2020}
\usepackage{times}
\usepackage{latexsym}

\usepackage{microtype}
\usepackage{graphics} 
\usepackage{graphicx} 
\usepackage{amsmath} 
\usepackage{amsfonts}
\usepackage{graphicx}
\usepackage{subcaption}
\usepackage[nopar]{lipsum}
\usepackage{enumitem} 
\setlist[itemize]{noitemsep, topsep=0pt}
\usepackage{comment} 
\usepackage{multirow}
\usepackage{rotating} 
\usepackage{xcolor}
\usepackage{colortbl}

\usepackage{url}
\usepackage{latexsym}
\aclfinalcopy

\title{PMIndia -- A Collection of Parallel Corpora of Languages of India}

\author{Barry Haddow and  Faheem Kirefu \\ School of Informatics \\ University of Edinburgh, \\ Edinburgh, Scotland \\ \texttt{\{bhaddow,fkirefu\}@inf.ed.ac.uk}}

\date{}

\begin{document}
\maketitle

\begin{abstract}
Parallel text is required for building high-quality machine translation (MT) systems, as well as for other multilingual NLP applications. For many South Asian languages, such data is in short supply. In this paper, we described a new publicly available corpus (PMIndia) consisting of parallel sentences which pair 13 major languages of India with English. The corpus includes up to 56000 sentences for each language pair. We explain how the corpus was constructed, including an assessment of two different automatic sentence alignment methods, and present some initial NMT results on the corpus.
\end{abstract}

\section{Introduction}
\label{sec:intro}

The languages of the South Asian subcontinent\footnote{Since the texts in this paper have been extracted from Indian sources, they include only languages spoken in India (and English) but many of these languages have significant communities (and official status) in other South Asian countries} have been poorly supported by parallel corpora and almost all would be considered ``under-resourced'' for machine translation. The largest parallel corpus we are aware of for  South Asian languages is the IIT Bombay English--Hindi corpus \cite{hindenglish}, containing about 1.5M sentence pairs from various sources. Parallel corpora for South Asian languages were released for the shared news translation tasks in English-Hindi and English-Gujarati at the WMT conference \cite{bojar-EtAl:2014:W14-33,barrault-etal-2019-findings}, as well as for the corpus-filtering task \cite{guzman-etal-2019-flores,koehn-etal-2019-findings}, which targeted English--Sinhala and English--Nepali.

The standard site for freely available parallel corpora is OPUS\footnote{\url{http://opus.nlpl.eu/index.php}} . However the coverage for the majority of South Asian languages is  limited, consisting mainly of religious texts and localisation corpora from the likes of  KDE, GNOME etc. These tend to be noisy and very domain-specific. More recently released corpora covering a wide variety of languages (including South Asian languages) are WikiMatrix extracted from Wikipedia \cite{2019arXiv190705791S}, and JW300 extracted from the Jehova's Witnesses website \cite{agic-vulic-2019-jw300}. 

In this paper we present a parallel corpus of languages of India (both Indo-Aryan and Dravidian) extracted from the website of the Prime Minister of India (\url{www.pmindia.gov.in}). This website publishes  background information, speeches and news from the Indian Prime Minister in
English and 13 languages of India (Assamese, Bengali, Gujarati, Hindi, Kannada, Malayalam, Manipuri, Marathi, Odia, 
Punjabi, Tamil, Telugu and Urdu). Almost all the  pages in the site are available both in English and Hindi, with translations into other languages made available to differing degrees. In this paper we focus on the collection of ``news updates'', since they provide a large and expanding store of multilingual articles. The website's policy\footnote{\url{https://www.pmindia.gov.in/en/website-policies/}} takes a permissive approach to copying, making the construction of our parallel corpus possible. We release our corpus under the CC-BY-4.0 licence.

The corpus is available from \url{http://data.statmt.org/pmindia} and the crawling code is available from \url{https://github.com/bhaddow/pmindia-crawler}.

\section{Languages of India}

India's linguistic diversity is said to be the second highest in the world (after Papua New Guinea), with 780 languages\footnote{\url{https://en.wikipedia.org/wiki/Languages_of_India}}. The vast majority of its languages fall into two language groups: the Indic, or Indo-Aryan, languages (a branch of the Indo-European family), and the Dravidian languages, an entirely separate family of 
languages spoken mainly in the south of India,  and also in  Sri Lanka. The Indian constitution lists 22 scheduled languages, all of which (apart from Sanskrit) have more than 2 million speakers, and 7 of which (according to the latest Ethnologue\footnote{\url{https://www.ethnologue.com/}}) are in the top 20
most spoken languages in the world. 

Languages in India are mostly written using one of a family of \emph{Brahmic scripts}. These scripts are all \emph{abugida} scripts, where each consonant-vowel sequence is written as a single unit, based on the consonant. Many of the languages of India have their own script, and whilst the form of the script is different, the phonemic inventories are almost the same (with the exception of the Tamil script, which has a significantly reduced  inventory), making automatic transliteration between Indian scripts quite feasible. A complication of processing  abugidas is that, because of the consonant-vowel sequence, each ``letter'' can be represented by up to three unicode characters, representing the consonants and the the vowel, written as a diacritic. A naive processing of
such a script can separate the bare consonant from its vowel diacritic, producing strange results. 

\section{Corpus Preparation}
\subsection{Crawling and Extraction}
\label{sec:prepare}
The news updates on the PMIndia website are presented in an ``infinite scroll'' format (\url{https://www.pmindia.gov.in/en/news-updates/}). Examination of the browser traffic showed that the requests for lists of URLs followed a fairly simple format, and so we were able to devise a custom scraper
to obtain all the news feed articles in the archive, for each language. Since there is a hyperlink in the html from each page to its corresponding English version, document alignment was straightforward. The local language versions of each article have translated language titles, but where no local language version of the article exists, the English language version is served instead. We were thus able to use the title
to determine whether we had successfully obtained an Indian language version of the article. The total article counts are shown in 
Table \ref{tab:articles}.
\begin{table}[ht]
    \centering
    \begin{tabular}{l|l||l|l}
       English (en)  & 5722  & Odia (or) & 3002 \\
       Hindi (hi) & 5244 &  Malayalam (ml) & 3407\\
       Bengali (bn) & 3937 &  Punjabi (pa) & 2764 \\
       Gujarati (gu) & 3872 &  Kannada (kn) & 2679\\
       Marathi (mr) & 3762 & Manipuri (mni) & 1612\\
       Telugu (te) & 3631 & Assamese (as) & 1571\\
       Tamil (ta) & 3606 &  Urdu  (ur) &  1413  \\
    \end{tabular}
    \caption{Counts of articles by language.}
    \label{tab:articles}
\end{table}

To extract the text of the articles from the html, we used the Alcazar\footnote{\url{https://github.com/saintamh/alcazar}} toolkit, which on inspection does a very good job of picking out the article body. The only additionaly processing we did on text extraction was removing the embedded tweets from the html prior to running Alcazar, since these are never translated, and interefered with our initial sentence alignment experiments.

To prepare the extracted text for sentence alignment, we split sentences using the Moses splitter \cite{koehn-etal-2007-moses}. We had to extend the Moses sentence splitter to support all the languages in PMIndia by (i) adding appropriate ``non-breaking prefixes'', such as the equivalents to ``Mr.'', and single letters; and (ii) adding support for beginning sentences with any of the character sets of India. We also modified the Moses splitter to better handle itemised lists, since these were a common occurance in the corpus; early experiments with alignment showed that inconsistencies in list handling were causing many errors.

\subsection{Sentence Alignment}
\label{sec:align}
We experimented with two different aligners; hunalign \cite{hunalign} and Vecalign \cite{thompson-koehn-2019-vecalign}. The former is based on length heuristics and a machine-readable bilingual dictionary (if available), whereas the latter uses sentence embeddings supplied by LASER \cite{2018arXiv181210464A} and a dynamic programming algorithm. In both cases, we only retained 1-1 alignments.

For hunalign, we used the crowd-sourced dictionaries from \cite{Pavlick-EtAl-2014:TACL}, arbitrarily selecting the first translation where there was more than one. There are dictionaries available for English to each of the other languages considered in this work, except for  Assamese, Manipuri, Odia and Urdu. 

Since Vecalign relies on LASER alignments, we were only able to apply it when these were available. They are available for English, and for 7 out of the 13 other languages (Bengali, Hindi, Malayalam, Marathi,  Tamil, Telugu and Urdu).

Where Vecalign is available, the final corpus was taken as the intersection of the corpora produced by hunalign and Vecalign. In other cases, we use just the alignments from hunalign. The total number of unique sentence pairs produced by each alignment method is shown in Table~\ref{tab:pair-counts}.


\begin{table}[ht]
    \centering
    \begin{tabular}{l|rrr}
     Pair   & hunalign & Vecalign & intersection  \\
     \hline
      as-en   & \textbf{9780} & -- & -- \\
      bn-en   & 40536 & 36278 & \textbf{29584} \\
      gu-en   & \textbf{50380} & --  & --  \\
      hi-en   & 62451 & 60081 & \textbf{56831} \\
      kn-en   & \textbf{35628} & --  & --  \\
      ml-en   & 42626 & 40435 & \textbf{33669}\\
      mni-en   & \textbf{7484} & --  & --  \\
      mr-en   & 43621 & 41959 &  \textbf{36131} \\
      or-en   & \textbf{38588} & --  & -- \\
      pa-en   & \textbf{34699} & --  & -- \\
      ta-en   & 47834 & 48923 & \textbf{39526} \\
      te-en   & 48962 & 48541 & \textbf{40283} \\
      ur-en   &  14617 & 13384 & \textbf{11167}\\
    \end{tabular}
    \caption{Counts of unique parallel sentence pairs, produced by the different align methods - hunalign, Vecalign, and the intersection of the two methods. Since Vecalign relies on LASER embeddings, it is not available for all languages. The released version is the intersection (where available) or the hunalign version (where intersect is not available). Counts in the released version are in bold.}
    \label{tab:pair-counts}
\end{table}

In order to provide an intrinsic assessment of the quality of the alignments, we first compared the Vecalign and hunalign alignments. Selecting the Vecalign alignments arbitrarily as the ``gold'', we calculated precision, recall and balanced f-score for
each of the 7 language pairs where we had alignments from both methods. The results are show in Table~\ref{tab:prf}.

\begin{table}[]
    \centering
    \begin{tabular}{l|ccc}
    Pair & precision & recall & f1 \\
    \hline
    bn-en &  .73 & .82  & .77 \\
    hi-en & .92 & .96 & .94 \\
    ml-en & .79 & .84 & .81\\
    mr-en &  .83 &  .87  & .85  \\
    ta-en & .83 & .82 & .83 \\
    te-en & .83 & .84 &  .83\\
    ur-en & .77 & .84  & .80 \\
    \end{tabular}
    \caption{Comparison of Vecalign and hunalign alignments. We take Vecalign as the ``gold'', and compute precision, recall and balanced f-score.}
    \label{tab:prf}
\end{table}

According to the results in Table~\ref{tab:prf}, the agreement between the two alignment methods is mostly around 80\%, and symmetric (in other words, the corpora produced by each method is about the same size). Better agreement is observed for hi-en, perhaps because the resources for that pair are of better quality, or perhaps because hi-en contains few ``indirect'' translations. By this we mean that some or most of the content was originally written in Hindi, then translated to English and the other languages of India. So there is a naturally ``looser'' alignment between the English texts and the non-Hindi languages of India. An analysis has not been done on which languages are sources and which are translationese, although this may be worth doing for the future.

We provide a further assessment of the quality of the alignment using human evaluation of part of the corpus. To do this, we selected the English-Tamil pair, recruited a native Tamil speaker who is fluent in English, and applied the 
KEOPS\footnote{\url{https://keops.prompsit.com/}} evaluation. In this tool, annotators are presented with proposed parallel sentence pairs from the corpus, and have to classify them into several different categories provided by the ELRC validation guidelines\footnote{\url{http://www.lr-coordination.eu/sites/default/files/common/Validation_guidelines_CEF-AT_v6.2_20180720.pdf}}.

For KEOPS-based evaluation, we randomly selected 3 different collections of 100 proposed sentence pairs. The first were pairs aligned by hunalign and not Vecalign, the second were pairs aligned by Vecalign but not hunalign, and the third were pairs aligned by both methods (i.e from the interesection). Only the last set of pairs is in the released corpous. The results are shown in Table \ref{tab:keops}.

We firstly note that a surprising number of sentences from the sample were classed as ``Wrong Tokenisation'' by the annotator. When asked about this, she explained that these were sentences where one language was missing information, as compared to the other. We hypothesise that this is either due to the translator deliberately removing output or adding explanations, or by inconsistent sentence splitting and alignment error linking a sentence to a partial translation. 

Looking at the sentence pairs picked out by only one of our automatic aligners, we see that a larger proportion
of the ``only hunalign'' pairs are valid translations than the ``only Vecalign'' pairs (41 vs. 26). We also note that Vecalign seems more inclined to select free translations (``Only vecalign''
has 13 free translations whilst ``Only hunalign'' has 2). 

For the sample picked out using 
the intersection of both aligners (the method we use for the final corpus) we see there is a high accuracy, with 79\% judged as valid alignments. In fact, if we adopt the more liberal measure of performance, considering that any pair not classed as Incorrect alignment or Wrong tokenisation is a correct alignment, then the performance is judged as 94\%. For the individual methods, taking into account the overlap measures in Table \ref{tab:prf}, we calculate the 
precision of Vecalign to be 70\% (conservative) and 88\% (liberal), compared to hunalign at 
72\% (conservative) and 88\% (liberal).



\begin{table}[ht]
    \centering
{\small    \begin{tabular}{l|p{1.5cm}p{1.5cm}p{1.3cm}}
    Category & Only Vecalign & Only hunalign & Both aligners \\
    \hline
    Valid translation & 26 & 41 & 79 \\
    Wrong language     & 0 & 0 & 0 \\
    Incorrect alignment     & 23 & 24 & 3 \\
    Wrong tokenisation & 18 &14  & 3 \\
    MT translation & 0&1 & 0 \\
    Translation error & 20 & 18  & 10 \\
    Free translation & 13 & 2 & 5 \\
    \end{tabular}
    \caption{Human comparison of alignment methods using KEOPS (see text) applied to 100 sample sentence pairs. We compare alignments only produced by hunalign, alignments only produced by Vecalign, and alignments from the intersection.}
    \label{tab:keops}
    }
\end{table}

\section{Machine Translation Experiments}
\label{sec:mt}
In order to provide a further validation of the corpus, and to give an indication of the state of automatic translation quality in each of the language pairs, we trained NMT systems for all pairs, in both directions. The systems are trained with the released versionsof the corpus for each language pair.

To prepare the data for NMT training, we randomly selected 1000 parallel sentences as a dev set and 1000 additional sentences for test, then preprocessed the data using the Moses \cite{koehn-etal-2007-moses} sequence of normalisation, tokenisation and truecasing. For the languages of India we used the Indic NLP toolkit\footnote{\url{https://github.com/anoopkunchukuttan/indic_nlp_library}} for tokenisation and normalisation, and omitted truecasing. We used separate BPE models \cite{sennrich-etal-2016-neural} on source and target to split the text into subwords, applying 10000 merges.

Our NMT model is a shallow Marian RNN, where we applied layer normalisation, label smoothing (0.1), exponential smoothing ($10^{-4}$) word dropout (0.1) and RNN dropout (0.2). We used a Marian working memory of 5GB, validating every 2000 updates, and stopping when cross-entropy failed to reduce for 10 consecutive validation points. It has recently been shown that performance on low-resource NMT is sensitive to hyperparameter tuning \cite{sennrich-zhang-2019-revisiting}, so we expect that better results could be obtained by tuning for each language pair individually. However our aim here is just to give a reasonable indication of how performance varies across the pairs, by choosing parameters generally appropriate for low-resource settings.

The MT results (\textsc{bleu} scores) are show in Table \ref{tab:nmt}. We evaluated on tokenized text, using the \texttt{multi-bleu.perl} script from Moses, since sacrebleu \cite{post1:2018:WMT} does not currently support South Asian languages.

\begin{table}[ht]
    \centering
    \begin{tabular}{l|rr}
    Language & en-X & X-en   \\
    \hline
      as   & 5.3 & 8.5\\
     bn     & 6.6 & 10.9\\
     gu  & 8.8 & 16.1\\
     hi    & 23.4 & 19.6\\
     kn     & 7.7 & 14.9\\
     ml      & 1.8 & 10.1\\
    mni      & 11.5& 13.9\\
    mr     & 6.0 & 12.4\\
    or    & 9.2 & 12.4\\
    pa     & 18.1 & 18.8\\
    ta  & 3.2 & 11.4\\
   te   & 7.4 & 14.3\\
   ur  & 16.5 & 15.3\\
    \end{tabular}
    \caption{Bleu scores for NMT systems built  on the PMIndia corpus, for English to/from languages of India, setting aside 1000 sentences for dev and 1000 for test. }
    \label{tab:nmt}
\end{table}

Looking at Table \ref{tab:nmt} we observer very low \textsc{bleu} scores for translation into languages of India, especially for the Dravidian languages (Kannada, Malayalam, Tamil and Telugu). The Dravidian languages are all agglutinative, making inward translation challenging, but also making evaluation of translation using word-based metrics like \textsc{bleu} less reliable. For the languages
with very small (less than 10k) data set sizes, the translation from/to Assamese shows poor results, as expected, but translation into
and out of Manipuri and Urdu scores much higher, despite the data set size. It's possible that the translation sample for these languages is more domain-specific than for other languages. For translation into English, the scores are nearly always higher, and sometimes much higher. This observation suggests that the low scores for translation into languages of India are mainly caused by the difficulties posed  by the languages themselves, and not by data set problems.

\section{Conclusions}
We have presented the PMIndia corpus, a parallel corpus including text from 13 South Asian paired with English, extracted from the Indian prime minister's website. All of these languages are low resource.

In the future we intend to release a multi-parallel version of the corpus, rather than the current ``English-centric'' version. Multilingual parallel corpora are even more scarce for South Asian languages, but exploiting multi-parallel corpora   in a one-to-many low resource setting yields considerable gains in \textsc{bleu} score \cite{dabre2019exploiting}.

\section*{Acknowledgments}

This work has received funding from the European Union’s Horizon 2020 research and innovation programme 
under grant agreement No 825299 (Gourmet). We thank the Prime Minister's Office of the Government of India for making the content
available for re-distribution.

\bibliography{biblio}
\bibliographystyle{acl_natbib}



\end{document}